\documentclass[runningheads]{llncs}

\usepackage{eccv}

\usepackage{eccvabbrv}
\usepackage{graphicx}
\usepackage{booktabs}
\usepackage{wrapfig}

\usepackage[accsupp]{axessibility}  
\usepackage{enumitem}
\usepackage{tabularx}
\usepackage{booktabs}
\usepackage{floatflt}

\usepackage[pagebackref,breaklinks,colorlinks,citecolor=eccvblue]{hyperref}

\usepackage{orcidlink}

\begin{document}

\title{
MaxMI: A Maximal Mutual Information Criterion for Manipulation Concept Discovery
}

\titlerunning{MaxMI}

\author{Pei Zhou \and Yanchao Yang}

\authorrunning{P.~Zhou et al.}

\institute{The University of Hong Kong \\
\email{pezhou@connect.hku.hk, yanchaoy@hku.hk}}

\newcommand{\ds}{\mathcal{D}}
\newcommand{\task}{\mathcal{T}}
\newcommand{\p}{\mathrm{p}}
\newcommand{\fcomp}{f^\mathrm{c}}
\newcommand{\smax}{\mathrm{Softmax}}
\newcommand{\mi}{\mathbb{I}}
\newcommand{\dt}{\Delta t}
\newcommand{\svar}{\boldsymbol{s}}
\newcommand{\loss}{\mathcal{L}}
\newcommand{\ent}{\mathbb{H}}

\maketitle

\begin{abstract}

We aim to discover manipulation concepts embedded in the unannotated demonstrations, which are recognized as key physical states. 
The discovered concepts can facilitate training manipulation policies and promote generalization. 
Current methods relying on multimodal foundation models for deriving key states usually lack accuracy and semantic consistency due to limited multimodal robot data. 
In contrast, we introduce an information-theoretic criterion to characterize the regularities that signify a set of physical states. 
We also develop a framework that trains a concept discovery network using this criterion, thus bypassing the dependence on human semantics and alleviating costly human labeling. 
The proposed criterion is based on the observation that key states, which deserve to be conceptualized, often admit more physical constraints than non-key states. 
This phenomenon can be formalized as maximizing the mutual information between the putative key state and its preceding state, i.e., Maximal Mutual Information (MaxMI). 
By employing MaxMI, the trained key state localization network can accurately identify states of sufficient physical significance, exhibiting reasonable semantic compatibility with human perception. 
Furthermore, the proposed framework produces key states that lead to concept-guided manipulation policies with higher success rates and better generalization in various robotic tasks compared to the baselines, verifying the effectiveness of the proposed criterion. 
Our source code can be found at \href{https://github.com/PeiZhou26/MaxMI}{https://github.com/PeiZhou26/MaxMI}.

\end{abstract}

\section{Introduction}\label{sec:intro}

\begin{figure}[!t]
    \centering
    \includegraphics[width=0.92\textwidth]{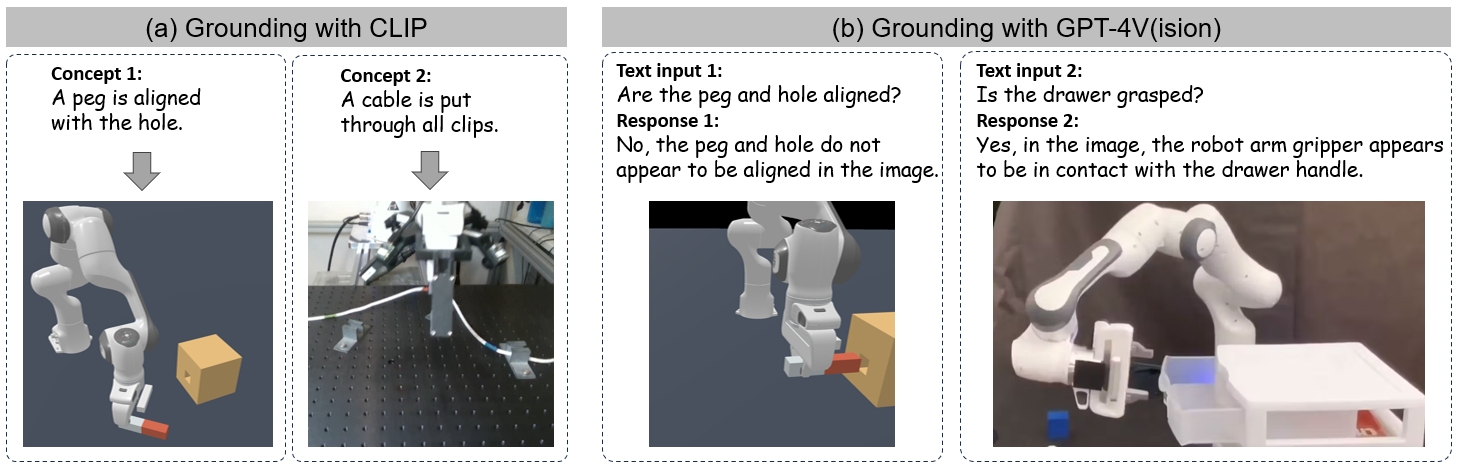}
    \caption{Grounding manipulation concepts via multimodal foundation models. (a): A manipulation concept is grounded using the multimodal encoders from CLIP~\cite{radford2learning} by checking the cosine similarity between the image features and the text embedding of the concept; (b): A multimodal LLM (GPT-4V) can also be used to ground a manipulation concept by directly asking it if the physical state presented renders the manipulation concept achieved. These examples demonstrate that concept grounding using large multimodal foundation models still lags due to the lack of robotic training data.}
    \label{fig:VLM-not-working}
\end{figure}

Recent advances in foundation models have shown great potential in promoting robot capabilities in perception~\cite{li2022grounded}, reasoning~\cite{brohan2023can}, and control~\cite{brohan2022rt}, among many other Embodied AI tasks~\cite{firoozi2023foundation}. 
Despite the remarkable ability to generalize in certain scenarios, their accuracy in grounding high-level manipulation concepts to low-level physical states is still limited due to the scarcity of annotated robot training data.
For example, 
when presenting images to multimodal foundation models (e.g., GPT-4V(ision)~\cite{openai2023gpt4}), 
and ask if a manipulation concept (e.g., ``the peg is aligned with the hole'') is achieved or not, 
we would usually obtain incorrect answers,
which signifies the incapability of current models 
in accurately understanding the physical conditions (Fig.~\ref{fig:VLM-not-working}).
The ability to faithfully tell whether a manipulation concept is realized given the low-level physical state is critical for the efficiency and safety of a robot learning system that integrates the foundation models.
Especially, given the many efforts that leverage foundation models to perform planning using high-level concepts in the form of step-by-step instructions~\cite{brohan2023can,chen2023autotamp} 
or to provide training signals in the form of value functions~\cite{di2023towards,huang2023voxposer}, 
all of which rely on an accurate connection (grounding) between the concepts described in language and the physical states.

One way to enhance the groundedness of foundation models is to collect more labeled training data, 
which inevitably induces a heavy annotation effort by asking humans to analyze the data and provide detailed descriptions of the key physical states, corresponding to manipulation concepts defined by human semantics. 
We refer to the grounding achieved from human annotations as top-down, since the reliance on high-level human understanding is assumed at the first place.
However, we propose that manipulation concepts exist on their own due to the significance in the corresponding physical states, but not the attached human semantics. 
In other words, we believe that {\it manipulation concepts are symbols of and derived from sets of physical states that possess certain regularities.}
For example, 
all the states that are instantiations of the manipulation concept ``turn on the faucet'' share the regularity or effect of running water, which is invariant of which language we use to describe the concept. 
If these key states are discovered, 
we can then assign them a description (one label) characterizing the set in the form of concept.
Accordingly, we denote the above grounding process as bottom-up, since a semantic name is only assigned after the concept is discovered from low-level states, which, in contrast, help reduce the annotation effort as required in the top-down grounding process.

In this work, we investigate the possibility of the aforementioned bottom-up concept grounding.
More explicitly, we aim for a learning framework that can automatically discover the key states in unannotated robot demonstrations, which deserve being assigned a concept name or description.
To achieve this goal, we propose a {\it Maximal Mutual Information} (MaxMI) criterion to characterize the regularities that endow a set of physical states with sufficient significance.
Specifically, the proposed MaxMI criterion measures the mutual information between a (key) state random variable (concept) and its preceding state variable (e.g., the low-level physical states can be treated as instantiations of the concept or state random variable).
Moreover, the proposed criterion promotes that the mutual information quantity should be a maximal when the discovered state variable signals a manipulation concept.
We further turn this metric into the training loss of a concept (key states) discovery network by leveraging a differentiable mutual information estimator.
After training, the discovery network receives observations from a demonstration and outputs the key states corresponding to a discovered concept.

Our experiments show that the concept discovery network trained with the MaxMI criterion can accurately localize physical states that align well with human semantics, 
alleviating the need for human supervision or heuristic rules. 
Moreover, we explicitly evaluate its efficacy by using discovered manipulation concepts to guide the training of manipulation policies. 
Our experiments verify that the discovered concepts effectively mitigate compounding errors in manipulation tasks, and the resulting policies outperform state-of-the-art methods in various robotic manipulation tasks. 
These results underscore the potential of the proposed MaxMI metric in self-supervised manipulation concept discovery and enhancing the grounding of manipulation concepts to the physical states.
{\bf To summarize,} 
this work contributes the following: 
1) An information-theoretic criterion named MaxMI that characterizes the significance of the physical states for manipulation concept discovery.
2) A framework that trains a neural network to discover manipulation concepts and accurately localize the corresponding physical states within a demonstration sequence.
3) A comprehensive evaluation of the proposed concept discovery framework with various ablations and for concept-guided manipulation policy learning.

\section{Related Work}\label{sec:related}

\noindent \textbf{Learning from Demonstrations.} 
Learning from Demonstration (LfD) refers to training robotic agents for complex tasks through expert demonstrations, which efficiently avoids costly self-exploration~\cite{schaal1996learning,ravichandar2020recent}. LfD methods include Inverse Reinforcement Learning (IRL), 
online Reinforcement Learning (RL) with demonstrations, and Behavior Cloning (BC)~\cite{argall2009survey}. 
IRL infers reward functions from observed behaviors, which could be computationally demanding~\cite{wulfmeier2015maximum,finn2016guided,zakka2022xirl}, 
while online RL with demonstrations combines dynamic online RL with offline guidance from the demonstrations~\cite{ho2016generative,riedmiller2018learning,stooke2021decoupling}. 
Further, BC falls into the regime of supervised learning to map input states to actions, 
but its performance is limited by the number of demonstrations and may subject to imitation errors due to overfitting~\cite{laskey2017dart,sasaki2020behavioral,zeng2021transporter,shridhar2023perceiver}. 
Our approach aims to minimize the imitation errors by leveraging key state guidance that can promote compositional generalization.

\noindent \textbf{Foundation Models for Task Planning.} Many recent works investigate the usage of foundation models~\cite{yang2023foundation}, such as Large Language Models (LLMs), for embodied planning, 
either in the form of code generation or by leveraging open-source tools for diverse interaction tasks~\cite{gupta2023visual,suris2023vipergpt,liu2023internchat,shen2023hugginggpt}. 
Besides virtual environment,
foundation models are also utilized to enable embodied agents to interact with the dynamic real-world environment~\cite{bommasani2021opportunities,fang2020learning}. 
Specifically,
foundation models can help convert language instructions and environmental information into control signals~\cite{huang2023instruct2act,driess2023palm}. 
Moreover, 
LLMs and Vision Language Models (VLMs) can be used to construct reward or value functions using carefully-crafted prompting strategies~\cite{yu2023language,huang2023voxposer,wang2023prompt}. 
However, all these models still lack the capability to accurately identify low-level physical states corresponding to high-level concepts due to the lack of annotated robot data. 

\noindent \textbf{Hierarchical Planning.} 
Hierarchical planning offers a structured approach to decomposing complex tasks into simpler, manageable sub-tasks across various abstraction levels, thereby facilitating the policy learning processes~\cite{hutsebaut2022hierarchical,jia2023chain,konidaris2012robot,stolle2002learning,liu2024infocon,weng2023towards}, 
particularly, with recent advancements that leverage the Chain of Thought (CoT) prompting technique~\cite{wei2022chain,wang2022self,yao2024tree}. 
One can provide manually decomposed high-level skills to facilitate learning with structures~\cite{sutton1999between,kulkarni2016hierarchical}, which may incur a heavy annotation burden.
There are also works studying unsupervised skill discovery, thus, eliminating the need for human annotations~\cite{bacon2017option,von2022self,kipf2019compile,yan2020self,nair2018visual}. 
Especially, AWE~\cite{shi2023waypointbased} extracts waypoints that can be used to linearly approximate a demonstration trajectory and leverages them to facilitate behavior cloning. 
Our method focuses on the abstraction of manipulation concepts through unsupervised discovery based on an information-theoretic metric, identifying highly informative states for guiding the policy learning.

\section{Method}\label{sec:method}

We aim to characterize the significance of a set of physical states, 
which shall enable self-supervised discovery of the states that are worth conceptualizing and can be used as guidance to facilitate the training of manipulation policies.
We first detail the problem setup 
and illustrate the key observations that motivate the Maximal Mutual Information (MaxMI) criterion.
We then describe a learning framework that leverages the proposed metric for discovering semantically meaningful manipulation concepts (or {\it key states}), 
as well as the policy training pipeline that serves as a testbed for the usefulness of the discovered concepts.

\subsection{Problem Statement}

Specifically, our goal is to develop a mechanism for localizing (key) physical states within a demonstration trajectory that are instantiations of a set of manipulation concepts, e.g., ``the peg is aligned with the hole.'' 
Formally, 
given a dataset of $N$ pre-collected demonstrations $\ds=\{\tau^i\}_{i=1}^N$ for a task $\task$, 
where each demonstration $\tau^i=\{(s^i_t,a^i_t)\}_{t=1}^{T_i}$ consists of a sequence of state-action pairs, 
we train a neural network to find the $N\times K$ temporal indices $\{t^i_{k}\}_{i,k=1}^{N,K}$, 
such that $s^i_{t^i_{k}}$ is the {\it key state} in trajectory $\tau^i$ corresponding to the $k$-th manipulation concept $\boldsymbol{\mu}_k$.
Next, 
we introduce the Maximal Mutual Information (MaxMI) criterion to enable the training without incurring costly manual annotations.

\subsection{Maximal Information at Key States}

\begin{figure*}[!t]
    \centering
    \includegraphics[width=0.9\textwidth]{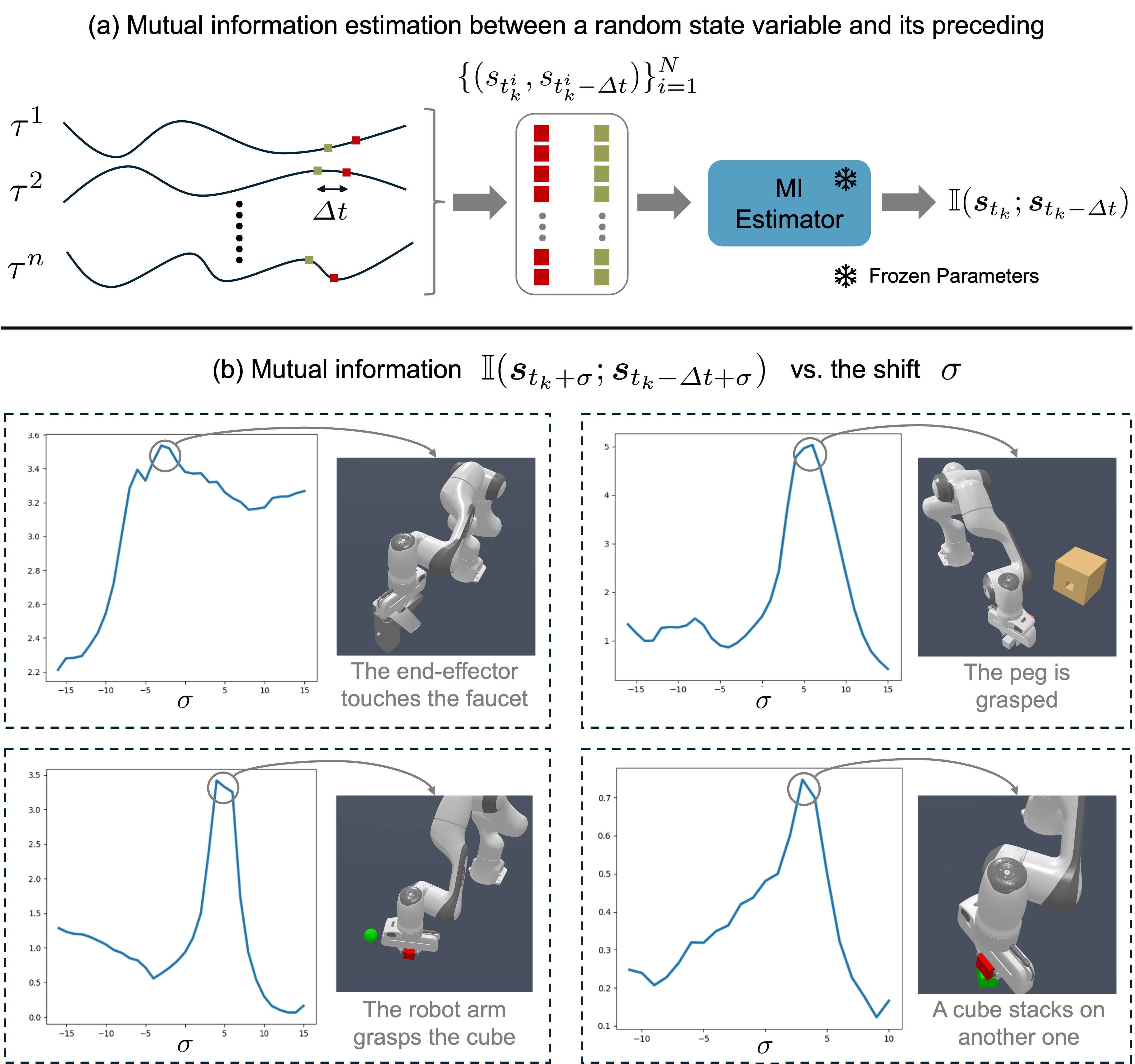}
    \caption{
    Mutual information between a random state variable and its preceding one achieves a maximum when the state variable coincides with a key state (manipulation concept), as verified across four manipulation tasks, namely, {\it Turn Faucet}, {\it Peg Insertion}, {\it Pick Cube}, and {\it Stack Cube}. The subfigures highlight the moments when the mutual information arrives at a peak, together with images that illustrate the corresponding key states.
    This phenomenon is commonly observed and suggests that one can discover manipulation concepts by maximizing such mutual information quantity.}
    \label{fig:mi_peak}
\end{figure*}

We propose that key states corresponding to manipulation concepts exist due to their physical significance but not human semantics, i.e., the state depicted by ``the peg is aligned with the hole'' is invariant to the language used to describe it (e.g., English or French).
In other words, 
we hypothesize that a key state is committed to more constraints than non-key states, which in turn endorses its physical significance.
Consequently, these constraints result in the fact that a key state should maximally inform its preceding states.
For example, 
the gripper of a robot can not be posed arbitrarily in the air
if the robot is to grasp the handle of a mug, making it easier to imagine what the preceding state looks like given the image (state) depicting ``handle grasped.''

More explicitly, 
we propose that the {\it mutual information between a key state and a preceding state should be maximal.}
To formalize, 
let $\svar_{t_k}$ be the random variable representing the $k$-th key state, 
such that a specific instantiation of the key state in trajectory $\tau^i$, i.e., $s_{t_k^i}$ (where the superscript $i$ is omitted for clarity),
can be treated as a sample from $\p(\svar_{t_k})$.
Similarly, let $\svar_{t_k-\dt}$ be the state preceding $\svar_{t_k}$ by a temporal interval $\dt>0$. 
Then the Shannon's mutual information $\mi(\svar_{t_k};\svar_{t_k-\dt})$ measures how much we know about $\svar_{t_k-\dt}$ after knowing $\svar_{t_k}$. 
Next, 
we plot $\mi(\svar_{t_k};\svar_{t_k-\dt})$ by shifting $t_k$ using the trajectories and ground-truth key state annotations (temporal indices) from \cite{jia2023chain}, 
to verify that the mutual information between a state and its preceding one achieves a maximal value at the key states.

{\bf Computing $\mi(\svar_{t_k};\svar_{t_k-\dt})$.}
Given the set of key states $\{s_{t^i_k}\}_{i=1}^N$ (manually annotated) corresponding to the concept $\boldsymbol{\mu}_k$ and the preceding states 
$\{s_{t^i_k-\dt}\}_{i=1}^N$,
we can compute the mutual information quantity $\mi(\svar_{t_k};\svar_{t_k-\dt})$ using the neural network proposed in \cite{hu2024infonet}, 
which leverages an attention architecture to predict the mutual information between two random variables from their samples.

We choose the neural estimator due to its efficiency, robustness, and differentiability, which is critical for training the proposed self-supervised concept discovery pipeline in the following.
Specifically, the neural network $\psi$ in \cite{hu2024infonet} takes in the paired sequences 
$\{(s_{t^i_k},s_{t^i_k-\dt})\}_{i=1}^N$,
and outputs the mutual information estimate according to the empirical joint distribution $\p(\svar_{t_k},\svar_{t_k-\dt})$ of the samples:
\begin{equation}
    \mi(\svar_{t_k};\svar_{t_k-\dt}) = \psi(\{(s_{t^i_k},s_{t^i_k-\dt})\}_{i=1}^N).
    \label{eq:mi-compute}
\end{equation}
With $\psi$, we can easily compute $\mi(\svar_{t_k+\sigma};\svar_{t_k-\dt+\sigma})$, $\sigma\in[-l,l]$.
In Fig.~\ref{fig:mi_peak}, 
we illustrate the procedure for computing $\mi(\svar_{t_k+\sigma};\svar_{t_k-\dt+\sigma})$, with $t_k$ and $\dt$ fixed (top), and display the resulted plots (bottom).
Our observations are:
\begin{itemize}
    \item[a.] The quantity $\mi(\svar_{t_k+\sigma};\svar_{t_k-\dt+\sigma})$ increases (under noise) when approaching a key state, i.e., $\sigma$ varies from $-l$ to $0$. 
    This aligns with our hypothesis that a key state is committed to more constraints than the non-key ones, thus, helps reduce the uncertainty of its preceding state, which in turn contributes to the increase in the mutual information.
    \item[b.] Moreover, $\mi(\svar_{t_k+\sigma};\svar_{t_k-\dt+\sigma})$ arrives at a maximal value around the key state ($\sigma=0$) and then starts to decrease, 
    since the constraints that help inform the preceding state are satisfied and then become ineffective when the key state is achieved. 
    For example, when the peg is grasped by the gripper, the gripper can move anywhere before the next key state ``the peg is aligned with the hole'' is in effect.
\end{itemize}

Therefore, the proposed hypothesis is confirmed with these observations and we can formalize a key state as the one that maximizes:
\begin{equation}
    \loss^{\mathrm{key}}(\svar_{t_k}) = \mi(\svar_{t_k};\svar_{t_k-\dt}),
    \label{eq:info-crit}
\end{equation}
which is functionally equivalent to the following quantity utilizing the neural estimator $\psi$:
\begin{equation}
    \loss^{\mathrm{key}}(\svar_{t_k}) = 
    \psi(\{(s_{t^i_k},s_{t^i_k-\dt})\}_{i=1}^N).
    \label{eq:infonet-con}
\end{equation}
We name the above as the {\bf Max}imal {\bf M}utual {\bf I}nformation (MaxMI) criterion of key states. 
We can then locate the instantiations of a key state, which satisfies the aforementioned observations, by maximizing the MaxMI metric. 
Next, we elaborate on the training objectives of the proposed self-supervised key state discovery framework.

\subsection{Key State Discovery}

\begin{figure}[!t]
    \centering
    \includegraphics[width=\textwidth]{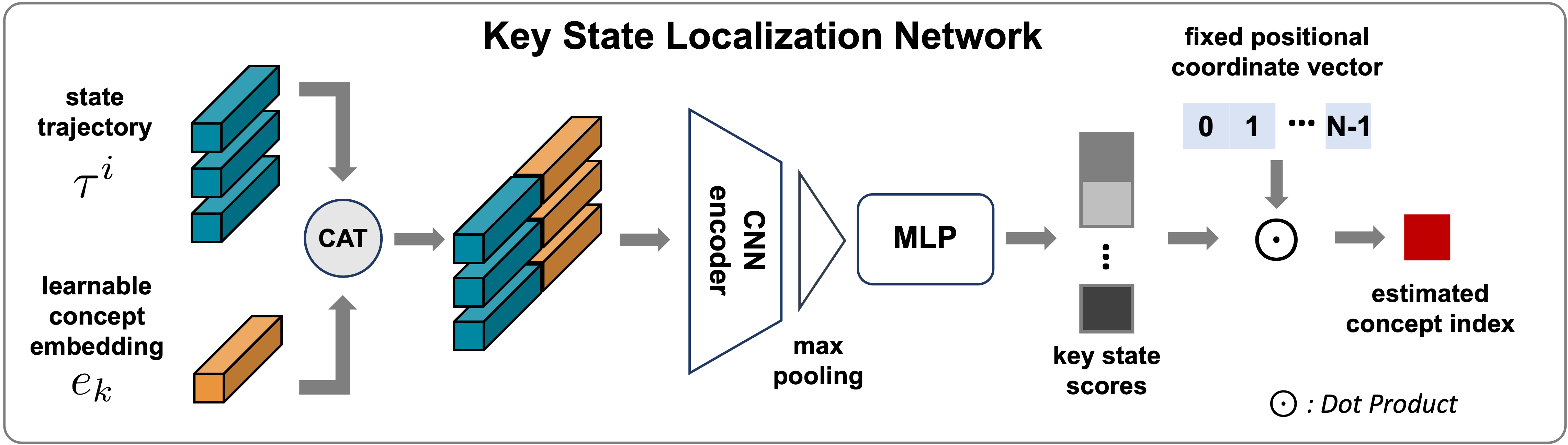}
    \caption{
    The proposed Key State Localization Network (KSL-Net) for manipulation concept discovery. 
    Every key concept (to be discovered and localized) is represented by a learnable embedding ($e_k$); 
    the concept embedding is then appended to all state vectors along a trajectory.
    These augmented state vectors are further processed by a fully convolutional encoder and a multi-layer perceptron (MLP) to derive the probability ($\p^i_k$) of each state being the identified key state.}
    \label{fig:framework}
\end{figure}

Discovering a key state now amounts to determining the temporal indices $\{t^i_k\}$ that maximizes the measure in Eq.~\ref{eq:infonet-con} by training a localization neural network (Fig.~\ref{fig:framework}).
We denote the localization network as $\phi$, 
which takes as input a trajectory $\tau^i=\{s_t^i\}_{t=1}^T, s_t^i\in\mathbb{R}^Q$ and a learnable concept embedding $e_k\in\mathbb{R}^P$ representing the key concept $\boldsymbol{\mu}_k$. 
We omit the actions in the trajectory, as by definition the key state should be sufficiently determined by the states themselves.
Also, we normalize each trajectory to the length of $T$ by interpolation to focus on the learning efficiency of the proposed MaxMI criterion instead of designing an autoregressive network architecture. 

The concept embedding $e_k$ is repeatedly concatenated with the states $s^i_t$, and the neural network $\phi$ maps the augmented physical states into a distribution corresponding to the key state selection probability, i.e., $\phi: \mathbb{R}^{(Q+P)\times T}\rightarrow\mathbb{R}^T$. 
We can also write $\p^i_k = \phi(\tau^i, \boldsymbol{\mu}_k)\in\mathbb{R}^T$ for the probability of each state in $\tau^i$ being the key state described by concept $\boldsymbol{\mu}_k$. 
The estimated key state location can then be obtained by applying an argmax over the predicted distribution $\p^i_k$.
However, due to the non-differentiability of the argmax function, we instead introduce a fixed temporal coordinate vector 
$\chi=[0,1,2,\ldots,T-1]$,
and compute the predicted key state location as the 
dot product between $\chi$ and the probability $\p^i_k$, e.g.,
$\hat{t}^i_k = \chi\cdot\p^i_k$, which makes the entire process differentiable.

The overall structure of the proposed Key State Localization Network (KSL-Net) is shown in Fig.~\ref{fig:framework}. 
A convolutional encoder is employed to fuse the states and concept embedding, 
followed by a max-pooling to aggregate information from various time-steps, 
and an MLP is used to output the key state selection probability.

{\bf Discovery objectives.}
As discussed, the predicted key state indices 
$\{\hat{t}^i_k\}_{i=1}^N$ should maximize the MaxMI criterion (Eq.~\ref{eq:infonet-con}) for a (to be discovered) concept $\boldsymbol{\mu}_k$.
However, the total number of concepts existing in the trajectories are not known beforehand.
To resolve this issue,
we assume that the maximum number of concepts for the manipulation tasks is $K$, e.g., 10, and train the KSL-Net $\phi$ by minimizing the following:
\begin{equation}
    \loss^{\mathrm{MaxMI}}(\phi;\ds) = -\sum_{k=1}^K\psi(
    \{s^i_{\hat{t}^i_k},s^i_{\hat{t}^i_k-\dt}\}_{i=1}^N),
    \label{eq:maxmi-obj}
\end{equation}
where the neural information estimator $\psi$ is pretrained and fixed \cite{hu2024infonet}.

However, training solely with Eq.~\ref{eq:maxmi-obj} would encourage all the discovered concepts to concentrate on one key state that has the largest mutual information value, even though there are many other (local) maximas that satisfy the maximal mutual information criterion.
To alleviate the clustering effect,
we propose a second regularization term that forces the discovered key states to be different by penalizing small distances between the localized indices of different key states (concepts):
\begin{equation}
    \loss^{\mathrm{div}}(\phi;\ds) = \sum_{i=1}^{N} \sum_{0<u<v<K} Sp(-|\hat{t}^i_u-\hat{t}^i_v|),
    \label{eq:regularization}
\end{equation}
where $|\cdot|$ is the $L$-1 norm, and $Sp$ is the softplus function: $ f(x) = \log(1 + \exp(x)) $.

The {\bf final loss} for training the key state localization network $\phi$ is:
\begin{equation}
    \loss^{\mathrm{discover}} 
    = \loss^{\mathrm{MaxMI}}(\phi;\ds) + \lambda\loss^{\mathrm{div}}(\phi;\ds),
\end{equation}
with $\lambda>0$ the weighting between the significance of the discovered concepts and their diversity.
Since K might be larger than the actual number of key states involved in a manipulation task,
we further apply non-maximum suppression to reduce the redundancy in the discovered concepts.
The quality of the discovered concepts (localized key states) is studied in the experiments.
Moreover, we evaluate the effectiveness of the proposed concept discovery framework (e.g., the MaxMI criterion) by training concept-guided manipulation policies using the localized key states.

\subsection{Manipulation Policy with Discovered Concepts}

We leverage the CoTPC framework \cite{jia2023chain} to learn concept-guided manipulation policy.
Specifically, the policy network predicts the action and estimates the key states (corresponding to different concepts) simultaneously during the training phase.
In this way, the concept guidance is injected into the action prediction process and can promote learning efficiency and generalization.
Instead of using manually annotated key states, we apply the localized key states from our concept discovery framework.
Given a trajectory $\tau^i=\{(s^i_t,a^i_t)\}_{t}$ and its localized key states $\{\hat{t}^i_k\}_k$, the CoTPC policy training loss is:
\begin{equation}
    \mathcal{L}^{\mathrm{policy}}=
    \sum_{i=1}^N\sum_t \|\bar{a}_t^i-a_t^i\|_2+ 
    \alpha \sum_{i=1}^N\sum_{k=0}^{K-1} \|\bar{s}_{\hat{t}^i_{k}}^{i}-s_{\hat{t}^i_{k}}^{i}\|_2,
    \label{eq:cotpc_loss}
\end{equation}
where $\bar{a}^i_t$ and $\bar{s}^i_{\hat{t}^i_k}$ are the predictions from the policy network introduced in \cite{jia2023chain}.
By optimizing Eq.~\ref{eq:cotpc_loss}, we now have a policy that can be applied to manipulation tasks, whose success rate in various settings shall indicate its performance as well as the usefulness of the discovered key states.

\section{Experiments}\label{sec:experiments}
In this section, we thoroughly evaluate the proposed key state discovery strategy. 
We first validate its effectiveness against baselines on complex multi-step tasks from ManiSkill2~\cite{gu2023maniskill2} and Franka Kithen~\cite{gupta2019relay,fu2020d4rl}, and then assess its performance and generalization in novel scenarios with unseen configurations and objects. Additionally, an ablation study is performed to explore various key state discovery approaches and the effect of the regularization term. 

\subsection{Experimental Setup}\label{exper_setup}
\noindent\textbf{Baselines.} 
The evaluation and comparison are performed with the following major baselines: Behavior Transformer (BeT)~\cite{shafiullah2022behavior}, Decision Diffuser (DD)~\cite{ajay2023is}, Decision Transformer (DT)~\cite{chen2021decision}, and Chain-of-Thought Predictive Control (CoTPC)~\cite{jia2023chain}. For more information about these methods, please refer to the appendix.

\begin{figure*}[!t]
    \centering
    \includegraphics[width=\textwidth]{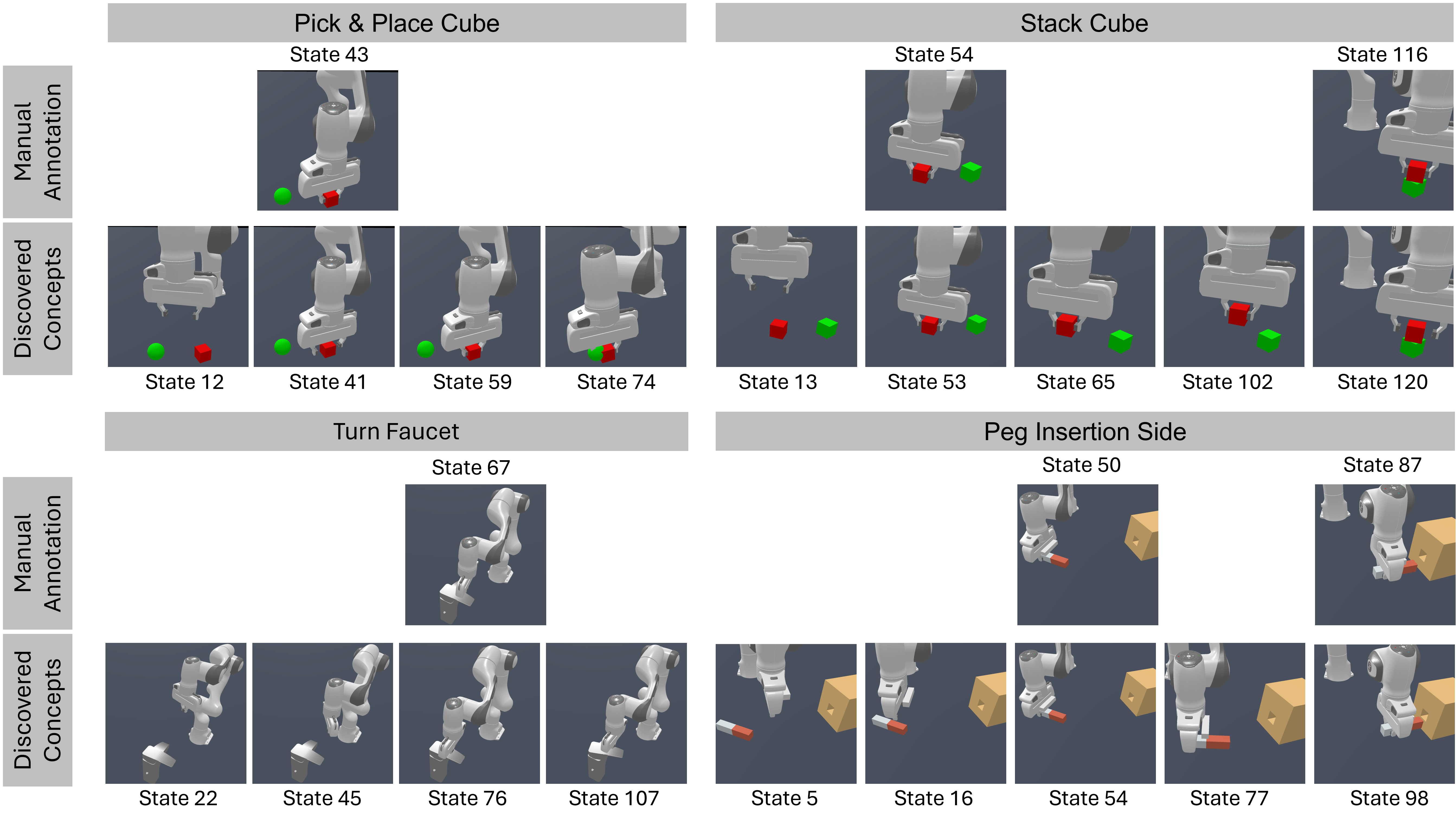}
    \caption{Examples of manually annotated key states and those discovered by the proposed pipeline, 
    across four distinct tasks: {\it Pick \& Place Cube}, {\it Stack Cube}, {\it Turn Faucet}, and {\it Peg Insertion Side}. 
    As observed, our method not only discovers the key states that align with human semantics, but also promotes more fine-grained manipulation concepts, which we show can effectively benefit the concept-guided policy learning.}
    \label{fig:dis_vs_man}
\end{figure*}

\noindent\textbf{Multi-Step Manipulation Tasks.}
We follow the experimental setting used in the most recent state of the 
art~\cite{jia2023chain}, and choose the same set of four complex multi-step object manipulation tasks from the Maniskill2 environments~\cite{gu2023maniskill2}. These four tasks are shown in Fig.~\ref{fig:dis_vs_man}. 
Please refer to the appendix for more details about the tasks.

\noindent\textbf{Experimental Setting.} We train the policies following the behavioral cloning (BC) paradigm without densely labeled rewards~\cite{jia2023chain}. 
Similar to CoTPC, we set the state space as the observation space, assuming full observability unless stated otherwise. For fair comparisons, we utilize the same set of 500 randomly sampled trajectories for a task. 
These trajectories, varying in initial states and environmental configurations, are sourced from multiple experts, added with diversity and randomness.
In the following sections, we detail the evaluation metric and experimental results.

\subsection{Discovered Concepts vs. Manually Annotated Concepts}

In Fig.~\ref{fig:dis_vs_man}, 
we compare manually annotated concepts with those discovered by our proposed method for four different tasks. 
Human annotations, restricted by their labor-intensive nature, 
are often sparse, thus providing limited information. 
Conversely, 
our method's flexibility in adjusting the number of key states enables the identification of a diverse range of concepts. 
This diversity, as evidenced in Fig.~\ref{fig:dis_vs_man}, 
underscores the capability of our method in enriching the concept guidance for more effective policy training.

On the other hand, 
Fig.~\ref{fig:dis_vs_man} shows that, 
even without explicitly injecting human semantics in the concept discovery process, 
the key concepts annotated by humans and a subset of the concepts discovered by our method exhibit a high degree of similarity. 
This suggests that our approach is able to discover concepts aligning with human understanding of the manipulation process. 
This alignment not only validates the efficacy of our concept discovery method but also highlights its potential in capturing intrinsic task-relevant features that resonate with human semantics, 
thereby enhancing the interpretability and applicability of our method in complex scenarios.

Also note that, 
since our approach allows a flexible selection of the number of discovered concepts,
more fine-grained concepts can be discovered that help promote the guided policy learning.
To prevent redundant manipulation concepts due to state noise, 
we consider concepts occurring closely in time as identical. 
Thus, we further employ a non-maximal suppression to prune repeatedly discovered concepts, with further details available in the appendix.

\begin{table*}[!t]
\caption{
Success Rates (SR) of the policies trained using our discovered concepts and various baselines across multiple tasks, 
evaluated with initial configurations seen during training. 
For more complex tasks like {\it Peg Insertion}, 
the success rates for intermediate sub-tasks such as ``grasp'' and ``align'' are also reported.}
\label{tab:seen_results}
\begin{center}
\begin{small}
\begin{sc}
\begin{tabular}{c c c c ccc c}
\toprule
 Methods & \multicolumn{1}{c}{P\&P} & \multicolumn{1}{c}{Stack} & \multicolumn{1}{c}{Turn} & \multicolumn{3}{c}{Peg} & mean \\
 & \multicolumn{1}{c}{Cube} & \multicolumn{1}{c}{Cube} & \multicolumn{1}{c}{Faucet} & \multicolumn{3}{c}{Insertion} & (\%) \\
\midrule
 & {\scriptsize Task SR} & {\scriptsize Task SR} & {\scriptsize Task SR} & {\scriptsize Grasp SR} & {\scriptsize Align SR} & {\scriptsize Insert SR} & {\scriptsize Task SR} \\
\cmidrule(lr){2-2}\cmidrule(lr){3-3}\cmidrule(lr){4-4}\cmidrule(lr){5-7}\cmidrule(lr){8-8}
BeT~\cite{shafiullah2022behavior} & 23.6 & 1.6 & 16.0 & 90.0 & 17.0 & 0.8 & 10.5 \\
DD~\cite{ajay2023is}  & 11.8 & 0.6 & 53.6 & 86.8 & 9.2 & 0.6 & 16.7 \\
DT~\cite{chen2021decision}  & 65.4 & 13.0 & 39.4 & 97.8 & 41.8 & 5.6 & 30.9 \\
MaskDP~\cite{liu2022masked}  & 54.7 &  7.8 &  28.8 & 62.6 & 5.8 & 0.0 & 22.8 \\
CoTPC~\cite{jia2023chain}  & \underline{75.2} & \underline{58.8} & \underline{56.4} & \underline{99.6} & \underline{98.2} & \underline{52.8} & \underline{60.8} \\
Ours  & \textbf{91.0} & \textbf{88.6} & \textbf{67.0} & \textbf{100.0} & \textbf{100.0} & \textbf{77.6} & \textbf{78.9} \\
\bottomrule
\end{tabular}
\end{sc}
\end{small}
\end{center}
\end{table*}

\subsection{Quantitative Results}

\noindent\textbf{Evaluation Metric.} 
Our primary evaluation metric is the task success rate. 
For complex multi-step tasks, 
we also report the success rate of completing intermediate sub-tasks, e.g., the Peg Insertion Side task, 
where the final objective is to insert the peg into a horizontal hole in the box. 
The intermediate sub-tasks include ``peg grasping by the robotic arm'' and ``alignment of the peg with the hole.'' 
During the evaluation phase, 
both seen (during training) and novel environmental configurations -- such as different initial joint positions of the robot -- are tested. 
For the Turn Faucet task, we also perform an additional evaluation on faucets with geometric structures not present in the training set.

\begin{table*}[t]
\caption{Success Rates (SR) of the policies trained using our discovered concepts and various baselines across multiple tasks, 
evaluated with initial configurations \textbf{unseen} during training. 
For more complex tasks like {\it Peg Insertion}, 
the success rates for intermediate sub-tasks such as ``grasp'' and ``align'' are also reported.
For the Turn Faucet task, we evaluate with faucets of novel geometric structures (zero-shot).}
\label{tab:unseen_results}
\begin{center}
\begin{small}
\begin{sc}
\begin{tabular}{c c c cc ccc}
\toprule
 Methods & \multicolumn{1}{c}{P\&P Cube} & \multicolumn{1}{c}{Stack Cube} & \multicolumn{2}{c}{Turn Faucet} & \multicolumn{3}{c}{Peg Insertion} \\
 & \multicolumn{1}{c}{(unseen)} & \multicolumn{1}{c}{(unseen)} & \multicolumn{2}{c}{(unseen \& 0-shot)} & \multicolumn{3}{c}{(0-shot)} \\
\midrule
 & {\scriptsize Task} & {\scriptsize Task} & {\scriptsize Task} & {\scriptsize \textit{Task}} & {\scriptsize \textit{Grasp}} & {\scriptsize \textit{Align}} & {\scriptsize \textit{Insert}}  \\
\cmidrule(lr){2-2}\cmidrule(lr){3-3}\cmidrule(lr){4-5}\cmidrule(lr){6-8}
DT~\cite{chen2021decision} & 50.0 & 7.0 & 32.0 & 9.0 & 92.3 & 21.8 & 2.0 \\
CoTPC~\cite{jia2023chain} & 70.0 & 46.0 & 57.0 & 31.0 & 95.3 & 72.3 & 16.8 \\
Ours & \textbf{76.0} & \textbf{63.0} & \textbf{58.0} & \textbf{35.0} & \textbf{98.3} & \textbf{81.8} & \textbf{21.3} \\
\bottomrule
\end{tabular}
\end{sc}
\end{small}
\end{center}
\end{table*}

\noindent\textbf{Main Results.} 
Tab.~\ref{tab:seen_results} and Tab.~\ref{tab:unseen_results} present the results on the seen and unseen environmental configurations (including zero-shot with novel objects), respectively. 
Tab.~\ref{tab:seen_results} demonstrates the superior performance of our proposed method in comparison to various baseline methods across multiple tasks.
It shows that the baselines have difficulties in handling complex multi-step tasks, 
while the policies trained with our discovered concepts consistently achieve the best performance, 
verifying the effectiveness of the proposed method in localizing physically meaningful key states. 
Due to the poor performance of baseline methods in dealing with novel configurations, 
we follow \cite{jia2023chain} to report the comparison with the two most effective baselines in unseen scenes. 
As evidenced in Tab.~\ref{tab:unseen_results}, our approach still outperforms these baselines, suggesting that our proposed key state discovery significantly enhances the generalization of the trained manipulation policies in new scenarios.

\begin{wraptable}[16]{r}{7cm}
\caption{
Evaluation in Franka Kitchen. 
The performance is measured as how many episodes out of 100 completed $k$, from 1 to 5, object-interaction tasks, higher the better; and ``MEAN'' denotes the average number of tasks completed per episode.}
\label{tab:franka_kit}
\begin{center}
\begin{small}
\begin{sc}
\begin{tabular}{ccccc}
\toprule
\# Tasks & MLP & DT~\cite{chen2021decision} & BeT~\cite{shafiullah2022behavior} & Ours \\
\midrule
1 & 72 & 100 & 100 & \textbf{100} \\
2 & 3 & 90 & 94 & \textbf{99} \\
3 & 2 & 74 & 71 & \textbf{85} \\
4 & 0 & 42 & 48 & \textbf{55} \\
5 & 0 & 5 & 4 & \textbf{7} \\
\midrule
mean & 0.77 & 3.11 & 3.17 & \textbf{3.46} \\
\bottomrule
\end{tabular}
\end{sc}
\end{small}
\end{center}
\end{wraptable}

\noindent\textbf{Franka Kitchen} 
To assess the efficacy in accomplishing long-horizon tasks, 
we train and test the concept-guided policies in the Franka Kitchen environment~\cite{gupta2019relay,fu2020d4rl}, 
where a 9-DoF Franka robot operates within a virtual kitchen. 
This environment features a total of seven object-interaction tasks. 
We utilize the dataset introduced by~\cite{shafiullah2022behavior} with 566 demonstrations. 
Each demonstration contains a sequence of four object-interaction tasks. 
The performance is measured by how many object-interaction tasks can be completed in one episode, i.e., 
the more tasks completed in one episode the better the policy. 
We check the ratio of the episodes that successfully achieve $\{1,2,3,4,5\}$ tasks out of 100 trials.
The results are reported in Tab.~\ref{tab:franka_kit}, 
with the maximum roll-out steps equal to 280. 
It shows that the discovered key concepts 
can effectively reduce the decision horizon, 
such that the policy trained with these concepts can achieve 
more subtasks within fixed steps in one roll-out.

\subsection{Ablation of Different Key State Selection Mechanisms}

\begin{table*}[!t]
\caption{Success rates of various key state selection strategies.}
\label{tab:key_selection}
\begin{center}
\begin{small}
\begin{sc}
\begin{tabular}{c cc cc cc cc }
\toprule
 Methods & \multicolumn{2}{c}{P\&P} & \multicolumn{2}{c}{Stack} & \multicolumn{2}{c}{Turn} & \multicolumn{2}{c}{Peg}\\
 & \multicolumn{2}{c}{Cube} & \multicolumn{2}{c}{Cube} & \multicolumn{2}{c}{Faucet} & \multicolumn{2}{c}{Insertion} \\
\midrule
 & {\scriptsize Seen} & {\scriptsize Unseen} & {\scriptsize Seen} & {\scriptsize Unseen} & {\scriptsize Seen} & {\scriptsize Unseen} & {\scriptsize Seen} & {\scriptsize Unseen} \\
\cmidrule(lr){2-3}\cmidrule(lr){4-5}\cmidrule(lr){6-7}\cmidrule(lr){8-9}
Last State & 66.6 & 60.0 & \underline{67.4} & 41.0 & 50.4 & 43.0 & 38.6 & 9.3 \\
\midrule
CLIP~\cite{di2023towards,radford2learning} & 73.2 & 65.0 & 66.0 & \underline{51.0} & 56.0 & 40.0 & 57.0 & 7.8 \\
\midrule
BLIP~\cite{li2022blip} & 69.0 & 63.0 & 52.8 & 26.0 & 53.8 & 44.0 & 52.6 & 11.3 \\
\midrule
BLIP2~\cite{li2023blip} & 73.0 & 59.0 & 56.6 & 43.0 & 54.6 & 51.0 & 59.4 & 11.3 \\
\midrule
FLAVA~\cite{singh2022flava} & 74.0 & 59.0 & 40.2 & 28.0 & 55.0 & 56.0 & 63.6 & \underline{13.0} \\
\midrule
CompILE~\cite{kipf2019compile} & 76.0 & 64.0 & 61.4 & 34.0 & \underline{61.8} & 48.0 & \underline{70.2} & 11.8 \\
\midrule
AWE~\cite{shi2023waypointbased} & \textbf{96.2} & \textbf{78.0} & 45.4 & 18.0 & 53.4 & \underline{57.0} & 31.6 & 4.3 \\
\midrule
Ours & \underline{91.0} & \underline{76.0} & \textbf{88.6} & \textbf{63.0} & \textbf{67.0} & \textbf{58.0} & \textbf{77.6} & \textbf{21.3} \\
\bottomrule
\end{tabular}
\end{sc}
\end{small}
\end{center}
\end{table*}

We further conduct an ablation with various key state selection strategies.   
The results are detailed in Tab.~\ref{tab:key_selection}. Please note that these policies are trained using the CoTPC framework~\cite{jia2023chain}.
The baselines mainly include those utilizing multimodal encoders, 
such as CLIP~\cite{radford2learning}, BLIP~\cite{li2023blip}, BLIP2~\cite{li2023blip}, and FLAVA~\cite{singh2022flava}, 
which ground concept descriptions (more details in the appendix) to the physical states via the feature similarity between multimodal encodings. 
We observe that the key states from VLMs are inaccurate and exhibit a lack of consistency across different trajectories (also refer to the appendix), resulting in the degenerated performance of the trained policies. 
Three other baselines include: ``last state,'' ComPILE~\cite{kipf2019compile}, and AWE~\cite{shi2023waypointbased}. 
The ``last state'' uses the end state as the only key state, 
whereas CompILE~\cite{kipf2019compile} learns for the subtask boundaries corresponding to structured behaviors; 
similarly, AWE~\cite{shi2023waypointbased} extracts waypoints as subtask boundaries that linearly approximate the trajectories. 
Again, the policies trained using our discovered concepts outperform by a significant margin,
verifying the effectiveness and usefulness of our key concept discovery method.

\subsection{Effect of the Regularization Term}

\begin{table*}[!t]
\caption{
Success rates of policies trained using key states discovered with different loss terms: ``Only PD'' stands for the pairwise distance regularization, ``Only MaxMI'' represents the information-theoretic criterion, and ``Ours'' is the proposed full model.}
\label{tab:criterion_selection}
\begin{center}
\begin{small}
\begin{sc}
\begin{tabular}{c cc cc cc cc }
\toprule
 Methods & \multicolumn{2}{c}{P\&P} & \multicolumn{2}{c}{Stack} & \multicolumn{2}{c}{Turn} & \multicolumn{2}{c}{Peg}\\
 & \multicolumn{2}{c}{Cube} & \multicolumn{2}{c}{Cube} & \multicolumn{2}{c}{Faucet} & \multicolumn{2}{c}{Insertion} \\
\midrule
 & {\scriptsize Seen} & {\scriptsize Unseen} & {\scriptsize Seen} & {\scriptsize Unseen} & {\scriptsize Seen} & {\scriptsize Unseen} & {\scriptsize Seen} & {\scriptsize Unseen} \\
\cmidrule(lr){2-3}\cmidrule(lr){4-5}\cmidrule(lr){6-7}\cmidrule(lr){8-9}
Only PD & 66.6 & 60.0 & 67.4 & 41.0 & 50.4 & 43.0 & 38.6 & 9.3 \\
\midrule
Only MaxMI & 73.2 & 65.0 & 66.0 & 51.0 & 56.0 & 40.0 & 57.0 & 7.8 \\
\midrule
Ours & \textbf{91.0} & \textbf{76.0} & \textbf{88.6} & \textbf{63.0} & \textbf{67.0} & \textbf{58.0} & \textbf{77.6} & \textbf{21.3} \\
\bottomrule
\end{tabular}
\end{sc}
\end{small}
\end{center}
\end{table*}
\begin{figure*}[!t]
    \centering
    \includegraphics[width=0.75\textwidth]{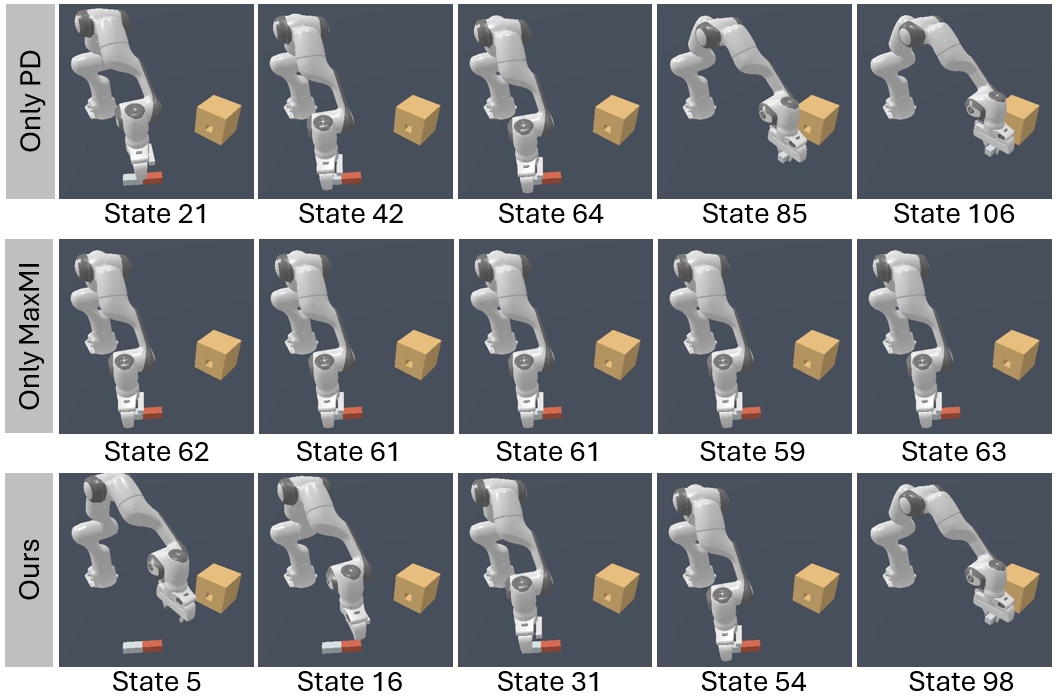}
    \caption{Key states discovered with different terms as discussed in Tab.~\ref{tab:criterion_selection}.}
    \label{fig:effect_pairwise}
\end{figure*}

Now we study the impact of the regularization term, e.g., the pairwise distance (PD) penalty.
The numerous results are presented in Tab.~\ref{tab:criterion_selection},
which validate the necessity of each term.
As shown in Fig.~\ref{fig:effect_pairwise}, 
key states in a demonstration discovered by different terms vary a lot. 
The pairwise distance term is observed to enforce diversity among the discovered concepts but alone can not ensure semantic meaningfulness, 
thereby simply exhibiting a uniform distribution across the trajectory.
In contrast, relying exclusively on the MaxMI criterion tends to discover clustered concepts around the state with the highest mutual information value, thus reducing the diversity of the concepts. 
The full model can discover more meaningful key states while maintaining diversity.

\section{Conclusion}\label{sec:conclusion}

We propose an information-theoretic metric that characterizes the physical significance of the key states in robotic manipulation tasks.
We further leverage the proposed metric to develop a self-supervised manipulation concept discovery pipeline that can produce meaningful key states. 
When used as guidance for training policies, these key states can lead to higher performance than a broad spectrum of baselines.
Additionally, we validate the necessity of the proposed terms through an extensive ablation study.
Our investigation also shows that the proposed MaxMI criterion alone may not guarantee the diversity of the discovered concepts due to its local modeling characteristic, which we deem a limit. 
We propose that future studies can resolve this issue by extending the metric to a multi-scale concept discovery framework so that global information on the trajectories can be accessed.

\subsection*{Acknowledgment}
This work is supported by 
the Early Career Scheme of the Research Grants Council (grant \# 27207224),
the HKU-100 Award, 
a donation from the Musketeers Foundation, 
the Microsoft Accelerate Foundation Models Research Program, 
and in part by the JC STEM Lab of Robotics for Soft Materials funded by The Hong Kong Jockey Club Charities Trust.
We also like to thank Zhengyang Hu for helping with the integration of InfoNet.

\bibliographystyle{splncs04}
\bibliography{main}

\begin{thebibliography}{10}
\providecommand{\url}[1]{\texttt{#1}}
\providecommand{\urlprefix}{URL }
\providecommand{\doi}[1]{https://doi.org/#1}

\bibitem{ajay2023is}
Ajay, A., Du, Y., Gupta, A., Tenenbaum, J.B., Jaakkola, T.S., Agrawal, P.: Is conditional generative modeling all you need for decision making? In: Proceedings of the International Conference on Learning Representations (ICLR) (2023)

\bibitem{argall2009survey}
Argall, B.D., Chernova, S., Veloso, M., Browning, B.: A survey of robot learning from demonstration. Robotics and autonomous systems  \textbf{57}(5),  469--483 (2009)

\bibitem{bacon2017option}
Bacon, P.L., Harb, J., Precup, D.: The option-critic architecture. In: Proceedings of the AAAI conference on artificial intelligence. vol.~31 (2017)

\bibitem{bommasani2021opportunities}
Bommasani, R., Hudson, D.A., Adeli, E., Altman, R., Arora, S., von Arx, S., Bernstein, M.S., Bohg, J., Bosselut, A., Brunskill, E., et~al.: On the opportunities and risks of foundation models. arXiv preprint arXiv:2108.07258  (2021)

\bibitem{brohan2022rt}
Brohan, A., Brown, N., Carbajal, J., Chebotar, Y., Dabis, J., Finn, C., Gopalakrishnan, K., Hausman, K., Herzog, A., Hsu, J., et~al.: Rt-1: Robotics transformer for real-world control at scale. arXiv preprint arXiv:2212.06817  (2022)

\bibitem{brohan2023can}
Brohan, A., Chebotar, Y., Finn, C., Hausman, K., Herzog, A., Ho, D., Ibarz, J., Irpan, A., Jang, E., Julian, R., et~al.: Do as i can, not as i say: Grounding language in robotic affordances. In: Conference on Robot Learning. pp. 287--318. PMLR (2023)

\bibitem{chen2021decision}
Chen, L., Lu, K., Rajeswaran, A., Lee, K., Grover, A., Laskin, M., Abbeel, P., Srinivas, A., Mordatch, I.: Decision transformer: Reinforcement learning via sequence modeling. Advances in neural information processing systems  \textbf{34},  15084--15097 (2021)

\bibitem{chen2023autotamp}
Chen, Y., Arkin, J., Zhang, Y., Roy, N., Fan, C.: Autotamp: Autoregressive task and motion planning with llms as translators and checkers. arXiv preprint arXiv:2306.06531  (2023)

\bibitem{di2023towards}
Di~Palo, N., Byravan, A., Hasenclever, L., Wulfmeier, M., Heess, N., Riedmiller, M.: Towards a unified agent with foundation models. In: Workshop on Reincarnating Reinforcement Learning at ICLR 2023 (2023)

\bibitem{driess2023palm}
Driess, D., Xia, F., Sajjadi, M.S., Lynch, C., Chowdhery, A., Ichter, B., Wahid, A., Tompson, J., Vuong, Q., Yu, T., et~al.: Palm-e: An embodied multimodal language model. arXiv preprint arXiv:2303.03378  (2023)

\bibitem{fang2020learning}
Fang, K., Zhu, Y., Garg, A., Kurenkov, A., Mehta, V., Fei-Fei, L., Savarese, S.: Learning task-oriented grasping for tool manipulation from simulated self-supervision. The International Journal of Robotics Research  \textbf{39}(2-3),  202--216 (2020)

\bibitem{finn2016guided}
Finn, C., Levine, S., Abbeel, P.: Guided cost learning: Deep inverse optimal control via policy optimization. In: International conference on machine learning. pp. 49--58. PMLR (2016)

\bibitem{firoozi2023foundation}
Firoozi, R., Tucker, J., Tian, S., Majumdar, A., Sun, J., Liu, W., Zhu, Y., Song, S., Kapoor, A., Hausman, K., et~al.: Foundation models in robotics: Applications, challenges, and the future. arXiv preprint arXiv:2312.07843  (2023)

\bibitem{fu2020d4rl}
Fu, J., Kumar, A., Nachum, O., Tucker, G., Levine, S.: D4rl: Datasets for deep data-driven reinforcement learning (2020)

\bibitem{gu2023maniskill2}
Gu, J., Xiang, F., Li, X., Ling, Z., Liu, X., Mu, T., Tang, Y., Tao, S., Wei, X., Yao, Y., Yuan, X., Xie, P., Huang, Z., Chen, R., Su, H.: Maniskill2: A unified benchmark for generalizable manipulation skills. In: International Conference on Learning Representations (2023)

\bibitem{gupta2019relay}
Gupta, A., Kumar, V., Lynch, C., Levine, S., Hausman, K.: Relay policy learning: Solving long-horizon tasks via imitation and reinforcement learning. arXiv preprint arXiv:1910.11956  (2019)

\bibitem{gupta2023visual}
Gupta, T., Kembhavi, A.: Visual programming: Compositional visual reasoning without training. In: Proceedings of the IEEE/CVF Conference on Computer Vision and Pattern Recognition. pp. 14953--14962 (2023)

\bibitem{von2022self}
von Hartz, J.O., Chisari, E., Welschehold, T., Valada, A.: Self-supervised learning of multi-object keypoints for robotic manipulation. arXiv preprint arXiv:2205.08316  (2022)

\bibitem{ho2016generative}
Ho, J., Ermon, S.: Generative adversarial imitation learning. Advances in neural information processing systems  \textbf{29} (2016)

\bibitem{hu2024infonet}
Hu, Z., Kang, S., Zeng, Q., Huang, K., Yang, Y.: Infonet: Neural estimation of mutual information without test-time optimization. arXiv preprint arXiv:2402.10158  (2024)

\bibitem{huang2023instruct2act}
Huang, S., Jiang, Z., Dong, H., Qiao, Y., Gao, P., Li, H.: Instruct2act: Mapping multi-modality instructions to robotic actions with large language model. arXiv preprint arXiv:2305.11176  (2023)

\bibitem{huang2023voxposer}
Huang, W., Wang, C., Zhang, R., Li, Y., Wu, J., Fei-Fei, L.: Voxposer: Composable 3d value maps for robotic manipulation with language models. arXiv preprint arXiv:2307.05973  (2023)

\bibitem{hutsebaut2022hierarchical}
Hutsebaut-Buysse, M., Mets, K., Latr{\'e}, S.: Hierarchical reinforcement learning: A survey and open research challenges. Machine Learning and Knowledge Extraction  \textbf{4}(1),  172--221 (2022)

\bibitem{jia2023chain}
Jia, Z., Liu, F., Thumuluri, V., Chen, L., Huang, Z., Su, H.: Chain-of-thought predictive control. arXiv preprint arXiv:2304.00776  (2023)

\bibitem{kipf2019compile}
Kipf, T., Li, Y., Dai, H., Zambaldi, V., Sanchez-Gonzalez, A., Grefenstette, E., Kohli, P., Battaglia, P.: Compile: Compositional imitation learning and execution. In: International Conference on Machine Learning. pp. 3418--3428. PMLR (2019)

\bibitem{konidaris2012robot}
Konidaris, G., Kuindersma, S., Grupen, R., Barto, A.: Robot learning from demonstration by constructing skill trees. The International Journal of Robotics Research  \textbf{31}(3),  360--375 (2012)

\bibitem{kulkarni2016hierarchical}
Kulkarni, T.D., Narasimhan, K., Saeedi, A., Tenenbaum, J.: Hierarchical deep reinforcement learning: Integrating temporal abstraction and intrinsic motivation. Advances in neural information processing systems  \textbf{29} (2016)

\bibitem{laskey2017dart}
Laskey, M., Lee, J., Fox, R., Dragan, A., Goldberg, K.: Dart: Noise injection for robust imitation learning. In: Conference on robot learning. pp. 143--156. PMLR (2017)

\bibitem{li2023blip}
Li, J., Li, D., Savarese, S., Hoi, S.: Blip-2: Bootstrapping language-image pre-training with frozen image encoders and large language models. arXiv preprint arXiv:2301.12597  (2023)

\bibitem{li2022blip}
Li, J., Li, D., Xiong, C., Hoi, S.: Blip: Bootstrapping language-image pre-training for unified vision-language understanding and generation. In: International Conference on Machine Learning. pp. 12888--12900. PMLR (2022)

\bibitem{li2022grounded}
Li, L.H., Zhang, P., Zhang, H., Yang, J., Li, C., Zhong, Y., Wang, L., Yuan, L., Zhang, L., Hwang, J.N., et~al.: Grounded language-image pre-training. In: Proceedings of the IEEE/CVF Conference on Computer Vision and Pattern Recognition. pp. 10965--10975 (2022)

\bibitem{liu2022masked}
Liu, F., Liu, H., Grover, A., Abbeel, P.: Masked autoencoding for scalable and generalizable decision making. Advances in Neural Information Processing Systems  \textbf{35},  12608--12618 (2022)

\bibitem{liu2024infocon}
Liu, R., Luo, Q., Yang, Y.: Infocon: Concept discovery with generative and discriminative informativeness. arXiv preprint arXiv:2404.10606  (2024)

\bibitem{liu2023internchat}
Liu, Z., He, Y., Wang, W., Wang, W., Wang, Y., Chen, S., Zhang, Q., Yang, Y., Li, Q., Yu, J., et~al.: Internchat: Solving vision-centric tasks by interacting with chatbots beyond language. arXiv preprint arXiv:2305.05662  (2023)

\bibitem{nair2018visual}
Nair, A.V., Pong, V., Dalal, M., Bahl, S., Lin, S., Levine, S.: Visual reinforcement learning with imagined goals. Advances in neural information processing systems  \textbf{31} (2018)

\bibitem{openai2023gpt4}
OpenAI: Gpt-4 technical report (2023)

\bibitem{radford2learning}
Radford, A., Kim, J.W., Hallacy, C., Ramesh, A., Goh, G., Agarwal, S., Sastry, G., Askell, A., Mishkin, P., Clark, J., et~al.: Learning transferable visual models from natural language supervision. In: International conference on machine learning. pp. 8748--8763. PMLR (2021)

\bibitem{ravichandar2020recent}
Ravichandar, H., Polydoros, A.S., Chernova, S., Billard, A.: Recent advances in robot learning from demonstration. Annual review of control, robotics, and autonomous systems  \textbf{3},  297--330 (2020)

\bibitem{riedmiller2018learning}
Riedmiller, M., Hafner, R., Lampe, T., Neunert, M., Degrave, J., Wiele, T., Mnih, V., Heess, N., Springenberg, J.T.: Learning by playing solving sparse reward tasks from scratch. In: International conference on machine learning. pp. 4344--4353. PMLR (2018)

\bibitem{sasaki2020behavioral}
Sasaki, F., Yamashina, R.: Behavioral cloning from noisy demonstrations. In: International Conference on Learning Representations (2020)

\bibitem{schaal1996learning}
Schaal, S.: Learning from demonstration. Advances in neural information processing systems  \textbf{9} (1996)

\bibitem{shafiullah2022behavior}
Shafiullah, N.M., Cui, Z., Altanzaya, A.A., Pinto, L.: Behavior transformers: Cloning $ k $ modes with one stone. Advances in neural information processing systems  \textbf{35},  22955--22968 (2022)

\bibitem{shen2023hugginggpt}
Shen, Y., Song, K., Tan, X., Li, D., Lu, W., Zhuang, Y.: Hugginggpt: Solving ai tasks with chatgpt and its friends in huggingface. arXiv preprint arXiv:2303.17580  (2023)

\bibitem{shi2023waypointbased}
Shi, L.X., Sharma, A., Zhao, T.Z., Finn, C.: Waypoint-based imitation learning for robotic manipulation. In: Conference on Robot Learning (2023)

\bibitem{shridhar2023perceiver}
Shridhar, M., Manuelli, L., Fox, D.: Perceiver-actor: A multi-task transformer for robotic manipulation. In: Conference on Robot Learning. pp. 785--799. PMLR (2023)

\bibitem{singh2022flava}
Singh, A., Hu, R., Goswami, V., Couairon, G., Galuba, W., Rohrbach, M., Kiela, D.: Flava: A foundational language and vision alignment model. In: Proceedings of the IEEE/CVF Conference on Computer Vision and Pattern Recognition. pp. 15638--15650 (2022)

\bibitem{stolle2002learning}
Stolle, M., Precup, D.: Learning options in reinforcement learning. In: Abstraction, Reformulation, and Approximation: 5th International Symposium, SARA 2002 Kananaskis, Alberta, Canada August 2--4, 2002 Proceedings 5. pp. 212--223. Springer (2002)

\bibitem{stooke2021decoupling}
Stooke, A., Lee, K., Abbeel, P., Laskin, M.: Decoupling representation learning from reinforcement learning. In: International Conference on Machine Learning. pp. 9870--9879. PMLR (2021)

\bibitem{suris2023vipergpt}
Sur{\'\i}s, D., Menon, S., Vondrick, C.: Vipergpt: Visual inference via python execution for reasoning. arXiv preprint arXiv:2303.08128  (2023)

\bibitem{sutton1999between}
Sutton, R.S., Precup, D., Singh, S.: Between mdps and semi-mdps: A framework for temporal abstraction in reinforcement learning. Artificial intelligence  \textbf{112}(1-2),  181--211 (1999)

\bibitem{wang2022self}
Wang, X., Wei, J., Schuurmans, D., Le, Q., Chi, E., Narang, S., Chowdhery, A., Zhou, D.: Self-consistency improves chain of thought reasoning in language models. arXiv preprint arXiv:2203.11171  (2022)

\bibitem{wang2023prompt}
Wang, Y.J., Zhang, B., Chen, J., Sreenath, K.: Prompt a robot to walk with large language models. arXiv preprint arXiv:2309.09969  (2023)

\bibitem{wei2022chain}
Wei, J., Wang, X., Schuurmans, D., Bosma, M., Xia, F., Chi, E., Le, Q.V., Zhou, D., et~al.: Chain-of-thought prompting elicits reasoning in large language models. Advances in Neural Information Processing Systems  \textbf{35},  24824--24837 (2022)

\bibitem{weng2023towards}
Weng, Y., Mo, K., Shi, R., Yang, Y., Guibas, L.: Towards learning geometric eigen-lengths crucial for fitting tasks. In: International Conference on Machine Learning. pp. 36958--36977. PMLR (2023)

\bibitem{wulfmeier2015maximum}
Wulfmeier, M., Ondruska, P., Posner, I.: Maximum entropy deep inverse reinforcement learning. arXiv preprint arXiv:1507.04888  (2015)

\bibitem{yan2020self}
Yan, M., Zhu, Y., Jin, N., Bohg, J.: Self-supervised learning of state estimation for manipulating deformable linear objects. IEEE robotics and automation letters  \textbf{5}(2),  2372--2379 (2020)

\bibitem{yang2023foundation}
Yang, S., Nachum, O., Du, Y., Wei, J., Abbeel, P., Schuurmans, D.: Foundation models for decision making: Problems, methods, and opportunities. arXiv preprint arXiv:2303.04129  (2023)

\bibitem{yao2024tree}
Yao, S., Yu, D., Zhao, J., Shafran, I., Griffiths, T., Cao, Y., Narasimhan, K.: Tree of thoughts: Deliberate problem solving with large language models. Advances in Neural Information Processing Systems  \textbf{36} (2024)

\bibitem{yu2023language}
Yu, W., Gileadi, N., Fu, C., Kirmani, S., Lee, K.H., Arenas, M.G., Chiang, H.T.L., Erez, T., Hasenclever, L., Humplik, J., et~al.: Language to rewards for robotic skill synthesis. arXiv preprint arXiv:2306.08647  (2023)

\bibitem{zakka2022xirl}
Zakka, K., Zeng, A., Florence, P., Tompson, J., Bohg, J., Dwibedi, D.: Xirl: Cross-embodiment inverse reinforcement learning. In: Conference on Robot Learning. pp. 537--546. PMLR (2022)

\bibitem{zeng2021transporter}
Zeng, A., Florence, P., Tompson, J., Welker, S., Chien, J., Attarian, M., Armstrong, T., Krasin, I., Duong, D., Sindhwani, V., et~al.: Transporter networks: Rearranging the visual world for robotic manipulation. In: Conference on Robot Learning. pp. 726--747. PMLR (2021)

\end{thebibliography}
\end{document}